\documentclass{article}

\usepackage{arxiv}

\usepackage[utf8]{inputenc} 
\usepackage[T1]{fontenc}    
\usepackage{hyperref}       
\usepackage{url}            
\usepackage{booktabs}       
\usepackage{amsfonts}       
\usepackage{nicefrac}       
\usepackage{microtype}      
\usepackage{lipsum}		
\usepackage{graphicx}
\usepackage{natbib}
\usepackage{doi}

\usepackage{bm}
\usepackage{amsmath, amssymb}
\usepackage{graphicx}
\usepackage{algorithm}
\usepackage{hyperref}
\usepackage{caption}
\usepackage{rotating}
\usepackage{multirow}
\usepackage{makecell}

\title{Least-Action-Guided Diffusion for Physical Extrapolation}

\author{%
	\href{https://orcid.org/0009-0003-7898-7511}{\includegraphics[scale=0.06]{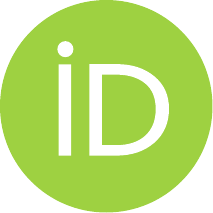}\hspace{1mm}Zhongxin Yang}$^{1}$\\
	$^{1}$College of Engineering, Peking University, Beijing 100871, China\\
	\And
	\href{https://orcid.org/0000-0001-7722-7885}{\includegraphics[scale=0.06]{orcid.pdf}\hspace{1mm}Yuanwei Bin}$^{2,3,5}$\thanks{Corresponding author}\\
	$^{2}$Ningbo Institute for Digital Twin, Eastern Institute of Technology, Ningbo 315200, Zhejiang, China\\
	$^{3}$Eastern Institute for Advanced Study, Eastern Institute of Technology, Ningbo 315200, Zhejiang, China\\
	$^{5}$Shenzhen Tenfong Technology Co., Ltd., Shenzhen 518000, Guangdong, China\\
	\texttt{ybin@eitech.edu.cn}\\
	\And
	\href{https://orcid.org/0000-0003-4940-5976}{\includegraphics[scale=0.06]{orcid.pdf}\hspace{1mm}Xiang I.~A. Yang}$^{4}$\\
	$^{4}$Mechanical Engineering, The Pennsylvania State University, University Park, PA 16802, USA\\
	\And
	\href{https://orcid.org/0000-0002-2913-4497}{\includegraphics[scale=0.06]{orcid.pdf}\hspace{1mm}Shiyi Chen}$^{2,3}$\\
	$^{2}$Ningbo Institute for Digital Twin, Eastern Institute of Technology, Ningbo 315200, Zhejiang, China\\
	$^{3}$Eastern Institute for Advanced Study, Eastern Institute of Technology, Ningbo 315200, Zhejiang, China\\
}

\renewcommand{\shorttitle}{Least-Action-Guided Diffusion for Physical Extrapolation}

\hypersetup{
pdftitle={Least-Action-Guided Diffusion for Physical Extrapolation},
pdfauthor={Yang et al.},
}

\begin{document}
\maketitle

\begin{abstract}		
Reliable extrapolation remains a central challenge for generative models in computational physics, because models trained over finite ranges of time, parameters, or geometries may produce physically inconsistent predictions outside the training distribution. 
We introduce a least-action-principle-guided diffusion, LAPG, a framework that promotes physical consistency during inference rather than relying solely on constraints imposed during training. 
The method combines a conditional score-based diffusion model with an action-derived physical guidance score. 
In the first stage, the learned score model generates an in-distribution proposal; in the second, an action-based variational prior refines this proposal toward the target out-of-distribution condition. 
This formulation turns the principle of least action into a differentiable inference-time correction mechanism and provides an alternative to pointwise residual penalties that often require empirical loss balancing.

We evaluate LAPG on representative ordinary- and partial-differential-equation systems, including free fall, conservative and dissipative spring--mass dynamics, interacting point vortices, and potential flow over parameterized airfoils. 
In temporal, parameter, and geometric extrapolation tests, LAPG reduces phase drift, preserves dissipative decay, captures vortex motion, and improves the lift response of airfoil flows compared with training-time physics-informed baselines.
\end{abstract}


\section{Introduction}
Machine learning is increasingly becoming a computational tool for modeling, prediction, and design in the physical sciences \cite{carleo2019machine,brunton2020machine,wang2023scientific}.
In computational physics, machine-learned models are often expected to serve not merely as interpolants of existing simulation data, but as efficient surrogates that can explore new parameter regimes, extend trajectories beyond observed time windows, and support design under changing physical conditions. 
This expectation imposes a stringent generalization requirement: useful models must remain physically reliable outside the distributions on which they were trained. However, such physical extrapolation remains a central difficulty for data-driven methods, especially when the target regime involves long-time evolution, unseen system parameters, or geometries absent from the training set.

Among recent generative approaches, diffusion models have emerged as particularly powerful tools for learning high-dimensional probability distributions \cite{sohl2015deep,ho2020denoising,song2019generative,song2020score}. 
By learning the score of progressively noise-perturbed data distributions, these models define samplers that have achieved strong performance in image synthesis and are now being adapted to scientific and engineering problems. 
In physical systems, diffusion models have been used to generate complex fields and trajectories, including turbulent flows, Lagrangian particle statistics, and spatiotemporal neural fields \cite{li2024synthetic,du2024conditional,gao2024generative}. 
These applications suggest that diffusion models can serve as data-driven generators for complex physical systems when repeated high-fidelity sampling is expensive.

Despite these advantages, diffusion models inherit a fundamental limitation of data-driven learning: the learned score is constrained primarily in the training distribution. 
When the target condition lies outside the training distribution, the reverse-time sampler follows a neural-network extrapolation of the learned score rather than a physical law. 
This difficulty is consistent with broader observations that neural networks can extrapolate unreliably outside the training distribution and that data-driven neural operators may suffer large errors when deployed beyond the support of the training set \cite{xu2021how,zhu2023reliable}. 
For physical systems, such errors are especially consequential: a generated sample may remain statistically plausible while developing phase drift in long-time trajectories \cite{linot2023stabilized}, incorrect amplitudes under parameter shifts \cite{zhu2023reliable}, violation of invariants or boundary conditions \cite{greydanus2019hamiltonian,krishnapriyan2021characterizing,bastek2025physics}, or distorted flow patterns for unseen geometries \cite{bhatnagar2019prediction}.
Thus, out-of-distribution (OOD) failure in physical generation is not only a loss of predictive accuracy, but also a loss of physical consistency during inference.

A major line of work addresses this issue by incorporating physical structure into learning. 
Physics-informed neural networks (PINNs) enforce governing equations by penalizing differential-equation residuals, initial conditions, and boundary conditions during training \cite{raissi2019physics}. 
Related approaches embed physical inductive biases more directly into the model architecture, for example through Hamiltonian or Lagrangian neural networks for dynamical systems \cite{greydanus2019hamiltonian,cranmer2020lagrangian}, symmetry- or equivariance-preserving networks \cite{satorras2021n,otto2023unified}, and physics-informed generative models that regularize diffusion training with physical residuals \cite{bastek2025physics}. 
These methods have significantly improved data efficiency and in-distribution physical fidelity, showing that physical knowledge can be a powerful constraint on learned models.

Nevertheless, most existing physics-informed strategies impose physical knowledge during model construction or training, often through soft penalty terms or architectural constraints \cite{raissi2019physics,karniadakis2021physics,wang2021understanding,bastek2025physics,cao2024parametric,cao2025rans}. 
After training, the model parameters are fixed, and extrapolative prediction still depends on how the learned map or score behaves outside the training domain. 
This motivates a complementary strategy: instead of enforcing physics only while learning the model, one may use physical principles directly during generation to guide each inference sample toward a physically consistent state.

In this work, we propose a least-action-principle-guided (LAPG) diffusion framework that enforces physical consistency at inference time. 
The method separates generation into two stages. 
First, a conditional score-based diffusion model is used to generate a physically plausible sample under an in-distribution condition. 
Second, the sample is refined toward the desired target condition by a physical guidance score. 
In this way, the learned score model provides a data-informed proposal, while the action-derived score supplies an inference-time correction that is not limited to the training distribution. 
The resulting sampler actively steers each generated trajectory or field toward physical consistency during generation, rather than relying solely on physical regularization imposed during training.

Although the least-action principle is most familiar in conservative Hamiltonian mechanics, the present method does not require the system to be conservative in this narrow sense. 
LAPG only requires a scalar variational functional whose stationary points or minimizers characterize physically admissible trajectories or fields. 
For dissipative dynamics, such functionals can be obtained by augmenting the conservative action with dissipation potentials, as in Rayleigh or Lagrange--d'Alembert formulations \cite{goldstein19801}. 
For fluid systems, action-like minimization principles have also been developed from variational formulations of vortex dynamics and from Gauss' principle of least constraint for incompressible flows \cite{khalifa2024vortex,taha2023minimization}. 
These examples allow the same inference-time guidance strategy to be applied to conservative, dissipative, and PDE-governed systems within a unified variational framework.

The contribution of this work is threefold. 
First, we define an action-residual score that can refine diffusion samples after the learned reverse process. 
Second, we apply the idea to phase-space trajectories and airfoil flow fields. 
Third, we evaluate the method under temporal, parameter, and geometry shifts and compare it with PINN-type baselines.

The remainder of this paper is organized as follows. 
Section~\ref{sec:methodology} introduces the LAPG formulation, including score-based diffusion and the action-derived physical prior.
Section~\ref{sec:experiments} describes the benchmark systems, diffusion model architecture, and baseline model. 
Section~\ref{sec:results} presents the extrapolation results. 
Section~\ref{sec:conclusion} summarizes the findings and discusses limitations and future extensions.

\section{Methodology}\label{sec:methodology}

We consider a family of physical systems specified by a condition vector $\bm c$, which may contain physical parameters, initial or boundary conditions, geometric representations, or mesh information. 
For each condition, the objective is to generate a physically admissible state ${\bf X}$. 
For the dynamical systems considered below, ${\bf X}$ denotes a discretized phase-space trajectory; for the airfoil-flow problems, it represents a discretized flow field. 
The training data are drawn from conditions $\bm c\in\mathcal{C}_{\rm train}$, while the target condition may lie outside this training distribution.

We build on score-based diffusion modeling \cite{song2020score}, where a forward stochastic process gradually perturbs data ${\bf X}_0$ into noise over a pseudo-time variable $\tau$. 
Given clean data ${\bf X}_0\sim p_0({\bf X}|\bm c)$, the forward process defines a family of perturbed conditional distributions $p_{\sigma_\tau}({\bf X}_\tau|\bm c)$ indexed by the noise level $\sigma_\tau \doteq \sigma(\tau)$.
The forward process is described by the stochastic differential equation (SDE):
\begin{equation}
	d{\bf X}_\tau
	=
	{\bf f}d\tau
	+
	g(\tau)d{\bf w}_\tau,
	\qquad \tau\in[0,T],
	\label{eq:forward_sde}
\end{equation}
where ${\bf f}$ is the drift coefficient, $g$ is the diffusion coefficient, and ${\bf w}_\tau$ is a standard Wiener process. 
In this work, we use the variance-exploding SDE (VESDE), for which ${\bf f}=0$ and
\begin{equation}
	g(\tau)
	=
	\sqrt{\frac{d\sigma^2(\tau)}{d\tau}},
	\qquad
	\sigma(\tau)
	=
	\sigma_{\min}
	\left(
	\frac{\sigma_{\max}}{\sigma_{\min}}
	\right)^{\tau/T}.
	\label{eq:ve_sde}
\end{equation}
Here, $T$ is the terminal diffusion time, while $\sigma_{\min}$ and $\sigma_{\max}$ are the minimum and maximum noise levels used in the forward noising process.
In all benchmarks the state variables are normalized before score-model training. We therefore use a fixed VESDE noise range, $\sigma_{\min}=0.01$ and $\sigma_{\max}=50.0$, for all systems, following the standard score-based diffusion setup \cite{song2020score}. These values were not tuned separately on OOD validation cases. The choice makes the smallest perturbation much smaller than the normalized data scale, while the largest perturbation is large enough that the terminal distribution is effectively Gaussian noise.
For this VESDE, the corresponding perturbation kernel is Gaussian,
\begin{equation}
	p({\bf X}_\tau|{\bf X}_0)
	=
	\mathcal{N}
	\left(
	{\bf X}_\tau;
	{\bf X}_0,
	\sigma^2(\tau){\bf I}
	\right),
	\label{eq:perturbation_kernel}
\end{equation}
so that a noisy sample can be written as ${\bf X}_\tau={\bf X}_0+\sigma(\tau)\bm z$, with $\bm z\sim\mathcal{N}({\bf 0},{\bf I})$.

The reverse process is governed by the corresponding score
\begin{equation}
	s({\bf X}_\tau, \sigma_\tau; \bm c)
	\equiv
	\nabla_{{\bf X}_\tau}\log p_{\sigma_\tau}({\bf X}_\tau|\bm c).
\end{equation}
This score is the gradient of the log-density of the perturbed conditional distribution. 
When known, the score provides the denoising direction of the perturbed data distribution.
Since this score is not available analytically for the physical datasets considered here, we approximate it by a neural network $S_{\theta}({\bf X}_\tau, \sigma_\tau;\bm c)$, where $\theta$ denotes the trainable network parameters, and train this network by denoising score matching:
\begin{equation}
	\mathcal{L}(\theta)
	=
	\mathbb{E}_{\tau,{\bf X}_0,\bm z}
	\left[
	\lambda(\tau)
	\left\|
	S_\theta({\bf X}_0+\sigma(\tau)\bm z,\sigma(\tau);\bm c)
	+
	\frac{\bm z}{\sigma(\tau)}
	\right\|_2^2
	\right].
	\label{eq:dsm_loss}
\end{equation}
For the VESDE, we use the standard weighting $\lambda(\tau)=\sigma^2(\tau)$. 
In Eq.~\eqref{eq:dsm_loss}, $\tau$ is sampled uniformly from $[0,T]$. Because $\sigma(\tau)$ follows the exponential VE schedule in Eq.~\eqref{eq:ve_sde}, uniform sampling in $\tau$ is equivalent to uniform sampling in $\log\sigma$ between $\log\sigma_{\min}$ and $\log\sigma_{\max}$. This schedule is used for all benchmarks.
After training, the optimized parameters are denoted by $\theta^\ast$.

At inference time, standard score-based generation integrates the reverse-time SDE from noise to data \cite{song2020score, anderson1982reverse}. 
In the present work, we augment this reverse process with an additional physical guidance term:
\begin{equation}
	d{\bf X}_\tau=
	[{\bf f}
	- g^2 S_{\theta^\ast}({\bf X}_\tau, \sigma_\tau;\bm c')]d\tau
	+ g\, d\bar{\bf w}_\tau
	\boxed{- \eta\, H(-\tau)\,
		\nabla_{{\bf X}_\tau}\log p_s({\bf X}_\tau|\bm c)\, d\tau}.
	\label{eq:sde_reverse}
\end{equation}
Here, $\eta$ controls the strength of the physical guidance. 
In our implementation, $\eta$ is not treated as an additional benchmark-dependent tuning parameter. 
After the physical prior is nondimensionalized, the product $\eta\Delta\tau$ in the refinement stage is absorbed into the optimizer learning rate used for physical refinement. 
The condition $\bm c$ denotes the desired target condition, which may lie outside the training distribution. 
Because the learned score model is trained only on $\mathcal{C}_{\rm train}$, we do not require it to extrapolate directly to $\bm c$. 
Instead, we evaluate the learned score at the closest in-distribution condition
\begin{equation}
	\bm c'
	=
	\operatorname*{argmin}_{\hat{\bm c}\in\mathcal{C}_{\rm train}}
	d(\bm c,\hat{\bm c}),
	\label{eq:closest_condition}
\end{equation}
where $d(\cdot,\cdot)$ denotes a distance metric in the normalized condition space; in this work we use the Euclidean distance, although other problem-specific metrics may also be used.

\begin{figure}
	\centering
	\includegraphics[width=0.7\linewidth]{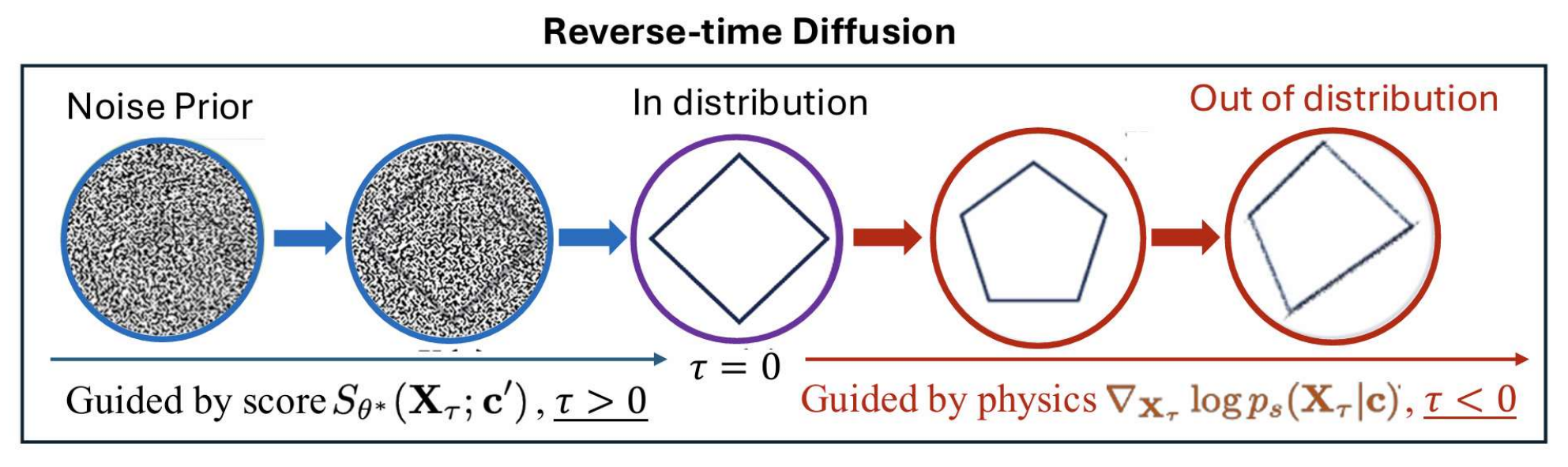}
	\captionsetup{type=figure}
	\captionof{figure}{\small Illustration of the reverse-time two-stage diffusion. Sampling starts from a noise prior and is guided by the learned score for $\tau>0$ to produce an in-distribution sample. At $\tau=0$, a physics guidance term, $\nabla_{{\bf X}_\tau}\log p_s({\bf X}_\tau|\bm c)$, is switched on and guides the dynamics for $\tau<0$ to refine the sample toward the target condition outside the training distribution.}
	\label{fig:fig1}
\end{figure}

The extension of the pseudo-time variable $\tau$ to negative values is an algorithmic device. 
It marks the transition from data-guided sampling to physics-guided refinement.
The Heaviside factor $H(-\tau)$ implements this switch in Eq.~\eqref{eq:sde_reverse}.
For $\tau>0$, the physical guidance is inactive and the sampler follows the learned score $S_{\theta^\ast}({\bf X}_\tau, \sigma_\tau;\bm c')$, producing a high-probability proposal under the training distribution. 
For $\tau\le 0$, the stochastic data-generation stage is followed by a physical refinement stage in which the action-derived score is activated. 
During this second stage, the noise term is removed and the update becomes an optimization of the generated state:
\begin{equation}
	{\bf X}_{\tau - \Delta \tau}
	=
	{\bf X}_{\tau}
	+
	(\eta \Delta \tau)
	\nabla_{{\bf X}_\tau}
	\log p_s({\bf X}_\tau|\bm c),
	\label{eq:physical_refinement_update}
\end{equation}
which gives the equivalent gradient-ascent direction on $\log p_s$.
The sign difference is due to the reverse-time convention. In Eq.~\eqref{eq:sde_reverse}, the sampler is integrated from larger $\tau$ toward smaller $\tau$, so the physical term $(\eta\Delta \tau)\nabla_{{\bf X}_\tau}\log p_s$ contributes a positive displacement along $\nabla_{{\bf X}_\tau}\log p_s$. After switching to the refinement stage, the objective is written as maximizing $\log p_s$. In the implementation, however, we do not use a single explicit gradient-descent step. Instead, the generated state is treated as the optimization variable and updated with a gradient-based optimizer such as Adam or stochastic gradient descent with momentum (SGDM). The optimizer, prescribed number of refinement iterations, and learning rate are specified in the experimental settings below. The runs use these prescribed iteration counts; during refinement, the action-variation residual is monitored and is regarded as converged when it no longer decreases appreciably.
Figure~\ref{fig:fig1} illustrates this two-stage reverse-time process.

The key component of LAPG is the physical prior $p_s({\bf X}_\tau|\bm c)$ in Eq.~\eqref{eq:sde_reverse}. 
We construct this prior from an action-based variational functional. 
Let $\mathcal{A}({\bf X};\bm c)$ denote an action or action-like functional associated with the physical system. 
A physically admissible trajectory or field corresponds to a stationary point, or in generalized formulations a minimizer, of this scalar functional. 
Thus, the variation of the action provides a global measure of physical inconsistency over the entire generated trajectory or field.

We convert this variational statement into a scalar ``unphysicality'' measure,
\begin{equation}
	U({\bf X}_\tau;\bm c)
	=
	\left(
	\frac{
		\widetilde{\delta \mathcal{A}}({\bf X}_\tau;\bm c)
	}
	{\sigma_{\mathcal{A}}}
	\right)^2,
	\label{eq:unphysicality}
\end{equation}
where $\widetilde{\delta \mathcal{A}}$ is a numerical approximation of the action variation and $\sigma_{\mathcal{A}}$ renders $U$ dimensionless. 
$\sigma_{\mathcal A}$ is not computed from the dataset and is not a tuned scale. It is introduced only as the reference unit of the action variation so that the residual entering $U$ is dimensionless.
Therefore, $\sigma_{\mathcal A}$ removes the physical units of $\widetilde{\delta\mathcal A}$ but does not further change its numerical value.
The physical prior is then defined as
\begin{equation}
	p_s({\bf X}_\tau|\bm c)
	\propto
	\exp[- U({\bf X}_\tau;\bm c)].
	\label{eq:physical_prior}
\end{equation}
States with small action variation are assigned high probability, whereas states with large action variation are exponentially suppressed. 
Taking the gradient of the log-prior yields the physical score
\begin{equation}
	\nabla_{{\bf X}_\tau}\log p_s({\bf X}_\tau|\bm c)
	=
	- \nabla_{{\bf X}_\tau} U({\bf X}_\tau;\bm c).
	\label{eq:physical_score}
\end{equation}
This score gives the update direction used to reduce the action variation of the generated sample.

The action variation is evaluated numerically during sampling. 
For trajectory problems, the generated state is the full discretized path ${\bf X}=\{\bm x_0,\ldots,\bm x_N\}$, where each $\bm x_i$ contains generalized coordinates and conjugate momenta. 
We introduce a finite set of virtual perturbations
\begin{equation}
	{\bf X}^{(a)} = {\bf X}+\epsilon\bm\xi^{(a)},
	\qquad a=1,\ldots,R,
	\label{eq:perturbation}
\end{equation}
where $\epsilon$ is a small scalar and $\bm\xi^{(a)}$ is an admissible perturbation direction. 
The quantity $\widetilde{\delta\mathcal A}$ used here is a multi-directional finite-difference stationarity residual, not the mathematical variation. A true stationary trajectory has zero first variation in every admissible direction; in practice we probe this condition with many prescribed admissible directions and penalize the mean squared directional residual. 
The perturbation of the generalized coordinates is chosen to vanish at the endpoints, consistent with the variational principle, while the perturbation of the conjugate momenta is chosen according to the degrees of freedom of each system. 
For each perturbation direction, the action variation is approximated by a finite difference,
\begin{equation}
	D_a\mathcal{A}({\bf X};\bm c)
	=
	\frac{
		\mathcal{A}({\bf X}+\epsilon\bm\xi^{(a)};\bm c)
		-
		\mathcal{A}({\bf X};\bm c)
	}
	{\epsilon}.
\end{equation}
The scalar residual used in Eq.~\eqref{eq:unphysicality} is the root-mean-square value over all probed directions,
\begin{equation}
	\widetilde{\delta \mathcal{A}}({\bf X};\bm c)
	=
	\left[
	\frac{1}{R}
	\sum_{a=1}^{R}
	\left(
	D_a\mathcal{A}({\bf X};\bm c)
	\right)^2
	\right]^{1/2}.
	\label{eq:action_variation_numeric}
\end{equation}
The finite differences in Eq.~\eqref{eq:action_variation_numeric} are used only to form the scalar multi-directional stationarity residual $\widetilde{\delta\mathcal A}$ and hence the scalar unphysicality $U$. 
The gradient in Eq.~\eqref{eq:physical_score} is then computed by automatic differentiation with respect to the generated state ${\bf X}_\tau$.
We use automatic differentiation because the discrete action evaluations are composed of differentiable tensor operations, so the required gradient can be obtained directly.
For the airfoil-flow problem, the same principle is applied to the variational degree of freedom controlling the potential-flow solution. 

In this construction, the learned score and the physical score play complementary roles. 
The learned score efficiently brings samples from noise to the neighborhood of the data manifold, while the action-derived score enforces the target physical condition during inference. 
The method therefore does not require retraining the diffusion model for each extrapolative target condition; instead, the action term is evaluated during generation.

\section{Experiments}\label{sec:experiments}

\begin{sidewaystable*}[p]
	\centering
	\caption{Benchmark systems, action functionals, training domains, and test conditions. Conditions in $\underline{\mathbf{bold}}$ indicate temporal, parameter, or geometric extrapolation outside the training domain. For potential flow past airfoils, the action functional and least-action formulation follow Ref.~\cite{taha2023minimization}. Incompressibility and boundary conditions are enforced by restricting the admissible flow field to be divergence-free and to satisfy no-penetration at solid surfaces.
		Quantities marked by $(\cdot)^*$ are nondimensionalized, and $[a,b]_n$ denotes $n$ samples over the interval $[a,b]$.
	}
	\label{tab:benchmark_setup}
	\scriptsize
	\setlength{\tabcolsep}{4pt}
	\renewcommand{\arraystretch}{1.25}
	\begin{tabular}{p{0.6cm}p{1.8cm}p{5.1cm}p{5.1cm}p{0.8cm}p{5.0cm}}
		\hline
		ID & System & Action $\mathcal{A}$ & Training domain & Case & Test condition \\ \hline
		
		\multirow{3}{*}{Q1} &
		\multirow{3}{*}{Free fall} &
		\multirow{3}{=}{$\displaystyle \int p\,dh-\left(\frac{p^2}{2m}+mgh\right)dt$} &
		\multirow{3}{=}{$m^*=1$; $g^*\in[5,15]_{1000}$; $t^*\in[0,2]_{128}$} &
		C1 & $m^*=1$, $g^*=10$, $t^*\in\underline{\mathbf{[0,4]}}$ \\
		& & & & C2 & $m^*=1$, $g^*=\underline{\mathbf{2}}$, $t^*\in\underline{\mathbf{[0,4]}}$ \\
		& & & & C3 & $m^*=1$, $g^*=\underline{\mathbf{30}}$, $t^*\in\underline{\mathbf{[0,4]}}$ \\ \hline
		
		\multirow{3}{*}{Q2} &
		\multirow{3}{*}{\parbox{1.8cm}{Undamped\\spring--mass}} &
		\multirow{3}{=}{$\displaystyle \int p\,dq-\left(\frac{p^2}{2m}+\frac{kq^2}{2}\right)dt$} &
		\multirow{3}{=}{$m^*=1$; $q_0^*,p_0^*\in[-1,1]_{20}$; $k^*\in[0.5,1.5]_{20}$; $t^*\in[0,T^*]_{128}$} &
		C1 & \mbox{$m^*=1$, $q_0^*=0$, $p_0^*=1$, $k^*=1$, $t^*\in\underline{\mathbf{[0,2T^*]}}$} \\
		& & & & C2 & \mbox{$m^*=1$, $q_0^*=0$, $p_0^*=1$, $k^*=\underline{\mathbf{3}}$, $t^*\in\underline{\mathbf{[0,2T^*]}}$} \\
		& & & & C3 & \mbox{$m^*=\underline{\mathbf{2}}$, $q_0^*=\underline{\mathbf{-2}}$, $p_0^*=1$, $k^*=1$, $t^*\in\underline{\mathbf{[0,2T^*]}}$} \\ \hline
		
		\multirow{3}{*}{Q3} &
		\multirow{3}{*}{\parbox{1.8cm}{Damped\\spring--mass}} &
		\multirow{3}{=}{\mbox{$\displaystyle \int p\,dq-\left(\frac{p^2}{2m}+\frac{kq^2}{2}+\int \mu\frac{p}{m}\,dq\right)dt$}} &
		\multirow{3}{=}{$m^*=1$; $q_0^*\in[0.5,2.0]_{20}$; $p_0^*=0$; $k^*\in[0.5,1.5]_{20}$; $\mu^*\in[1.5,3.0]_{20}$; $t^*\in[0,50]_{2048}$} &
		C1 & \mbox{$m^*=1.0$, $q_0^*=1$, $p_0^*=0$, $k^*=1.0$, $\mu^*=\underline{\mathbf{6.0}}$, $t^*\in[0,50]$} \\
		& & & & C2 & \mbox{$m^*=\underline{\mathbf{0.1}}$, $q_0^*=1$, $p_0^*=0$, $k^*=\underline{\mathbf{0.1}}$, $\mu^*=\underline{\mathbf{0.1}}$, $t^*\in[0,50]$} \\
		& & & & C3 & \mbox{$m^*=\underline{\mathbf{0.5}}$, $q_0^*=\underline{\mathbf{10}}$, $p_0^*=0$, $k^*=1.0$, $\mu^*=2.0$, $t^*\in[0,50]$} \\ \hline
		
		\multirow{3}{*}{Q4} &
		\multirow{3}{*}{\parbox{1.8cm}{Point\\vortices}} &
		\multirow{3}{=}{$\displaystyle \int \Gamma_1 y_1\,dx_1+\Gamma_2 y_2\,dx_2
			+\int\frac{\Gamma_1\Gamma_2}{2\pi}\ln|\bm r_1-\bm r_2|\,dt$} &
		\multirow{3}{=}{$x_{2,0}^*\in[0.5,2.0]_{50}$; $\Gamma_2^*\in[-0.5,2.0]_{40}$; $t^*\in[0,T^*]_{128}$} &
		C1 & $x_{2,0}^*=1$, $\Gamma_2^*=1$, $t^*\in\underline{\mathbf{[0,2T^*]}}$ \\
		& & & & C2 & $x_{2,0}^*=\underline{\mathbf{4}}$, $\Gamma_2^*=1$, $t^*\in\underline{\mathbf{[0,2T^*]}}$ \\
		& & & & C3 & $x_{2,0}^*=1$, $\Gamma_2^*=\underline{\mathbf{4}}$, $t^*\in\underline{\mathbf{[0,2T^*]}}$ \\ \hline
		
		\multirow{3}{*}{Q5} &
		\multirow{3}{*}{Airfoil} &
		\multirow{3}{=}{$\displaystyle \frac{1}{2}\int\rho\left(\partial_t\bm u+\bm u\cdot\nabla\bm u\right)^2d\bm x$} &
		\multirow{3}{=}{$\alpha^\ast\in[-1,0]_{30}$; $\beta^\ast=0$; $R^\ast\in[1.02,2.02]_{30}$; ${\rm AoA}\in[0^\circ,10^\circ]_{11}$} &
		C1 & \mbox{$\alpha^*=-0.172$, $\beta^*=0.0$, $R^*=1.27$, ${\rm AoA}=0^\circ$} \\
		& & & & C2 & \mbox{$\alpha^*=-0.172$, $\beta^*=0.0$, $R^*=1.27$, ${\rm AoA}=\underline{\mathbf{30^\circ}}$} \\
		& & & & C3 & \mbox{$\alpha^*=-0.150$, $\beta^*=\underline{\mathbf{0.1}}$, $R^*=1.20$, ${\rm AoA}=\underline{\mathbf{30^\circ}}$} \\ \hline
		
	\end{tabular}
\end{sidewaystable*}

\subsection{Benchmark systems}
We evaluate LAPG on five benchmark systems, denoted Q1--Q5, spanning conservative dynamics, dissipative dynamics, vortex dynamics, and airfoil-flow field generation. 
Although the first four systems are low-dimensional, they are deliberately chosen as controlled probes of physical extrapolation. 
Their analytical reference solutions allow unambiguous error evaluation, systematic separation of temporal and parameter extrapolation, and diagnosis of specific failure modes such as phase drift, amplitude error, incorrect dissipative decay, and violation of orbital structure. 
Such canonical dynamical systems are standard test beds for structure-preserving numerical methods and physics-informed learning because they expose long-time stability and physical-consistency errors that may be obscured in larger simulations \cite{greydanus2019hamiltonian,cranmer2020lagrangian,goldstein19801,hairer2006geometric}. 
The fifth system, potential flow over Joukowsky airfoils, extends the evaluation to a spatially distributed field problem with geometric and condition extrapolation.

Q1 is free fall under a constant gravitational field. 
The coordinate is the vertical position $h$, the momentum is $p=m\dot h$, $m$ is the mass, and $g$ is the gravitational acceleration. 
This case tests whether LAPG can extrapolate both in time and in the gravity $g$.

Q2 is an undamped spring--mass oscillator. 
The coordinate $q$ denotes displacement, $p=m\dot q$ is the momentum, $m$ is the mass, and $k$ is the spring stiffness. 
The initial displacement and momentum are denoted by $q_0$ and $p_0$, respectively. 
The characteristic period is $T=2\pi\sqrt{m/k}$. 
This conservative system tests long-time phase accuracy and parameter extrapolation in $m$ and $k$.

Q3 is a damped spring--mass system. 
In addition to $m$, $k$, $q_0$, and $p_0$, the system contains a linear damping coefficient $\mu$. 
This benchmark introduces dissipation and tests whether the generated trajectory preserves the correct decay behavior when the mass, stiffness, or damping coefficient lies outside the training range.

Q4 is a two-point-vortex system. 
The vortex positions are $\bm r_i=(x_i,y_i)$ and the vortex circulations are $\Gamma_i$, with $i=1,2$. 
Without loss of generality, the first vortex, with circulation $\Gamma_1$, is fixed initially at $\bm r_1(0)=(0,0)$, while the initial position of the second vortex is parameterized by $x_{2,0}$ and its circulation by $\Gamma_2$. 
This case tests extrapolation in interacting Hamiltonian dynamics, including changes in initial separation and circulation ratio. 
For Q1--Q4, we represent the generated state in phase space. 
Here, $q$ denotes a generalized coordinate and $p$ denotes its conjugate momentum; the generated state is a discretized trajectory ${\bf X}=\{\bm x_i\}_{i=0}^{N}$, where $\bm x_i=[\bm q_i,\bm p_i]^T$. 
This representation is natural for action-based guidance because the action can be written in canonical form using the phase-space path. 
It also reduces the burden of temporal extrapolation. 
A diffusion model trained on fixed-window time trajectories has no direct mechanism to change the terminal time associated with an unseen horizon, because its samples are tied to the temporal grid used during training. 
In the phase-space formulation, the generated path is instead treated as an ordered geometric path in phase space rather than as a fixed time-indexed sequence.
For Q1--Q4, time reconstruction uses only the coordinate displacement and the corresponding coordinate velocity, not the full phase-space velocity.
\[
\bm v_q({\bf x}_{i+1/2};\bm c)
=
\frac{\partial\mathcal H}{\partial\bm p}({\bf x}_{i+1/2};\bm c),
\qquad
{\bf x}_{i+1/2}=\frac{{\bf x}_i+{\bf x}_{i+1}}{2}.
\]
The segment time is computed componentwise from displacement divided by velocity and then averaged,
\[
\Delta t_i
\approx
\frac{1}{d_q}\sum_{j=1}^{d_q}
\frac{\Delta q_{i,j}}
{v_{q,j}({\bf x}_{i+1/2};\bm c)} ,
\]
where $d_q$ is the number of generalized-coordinate components in $\bm q$.
Extreme component-wise values relative to the expected time step are clipped for numerical stability. 
The reconstructed time is $t_0=0$ and $t_n=\sum_{i=0}^{n-1}\Delta t_i$, and the resulting trajectory is finally interpolated to a uniform time grid.

For temporal extrapolation, the target horizon is imposed through the terminal constraint of the phase-space path. 
In implementation, this constraint is represented as a terminal hypersurface rather than as a fully specified endpoint: finite entries in $q_{\texttt{end}}$ or $p_{\texttt{end}}$ are fixed.
For example, one can fix a terminal coordinate such as $q_N=q_{\texttt{end}}$ while leaving the terminal momentum unconstrained; the refinement then adjusts the full path subject to this endpoint hypersurface.
\begin{figure}
	\centering
	\includegraphics[width=0.31\linewidth]{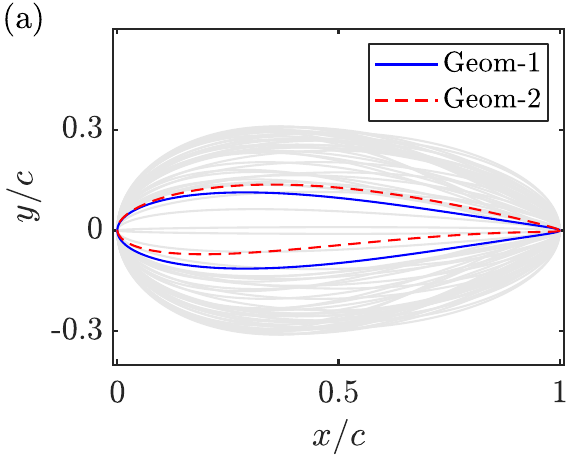}
	\includegraphics[width=0.31\linewidth]{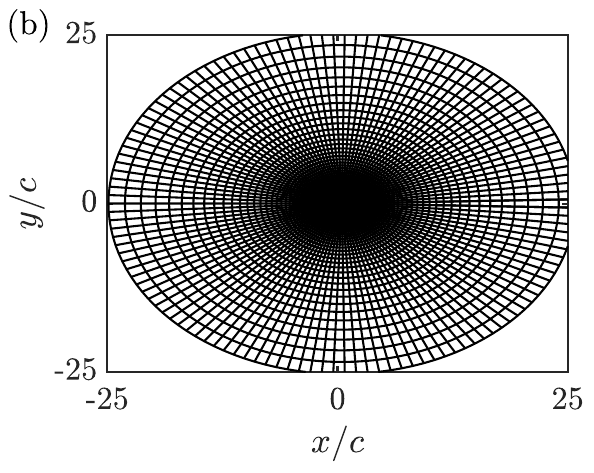}
	\includegraphics[width=0.31\linewidth]{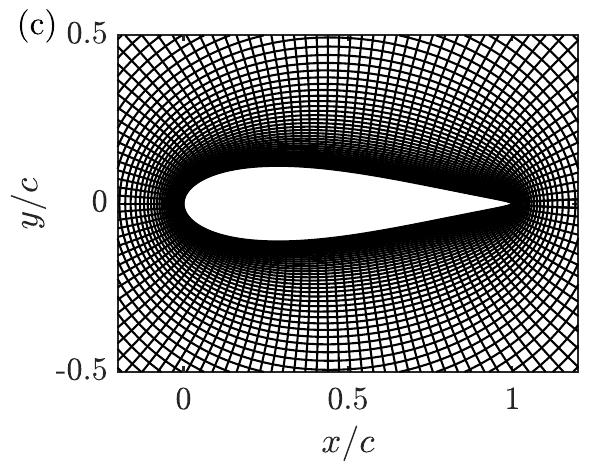}
	\caption{Airfoil dataset and mesh. (a) Representative Joukowsky airfoil geometries. ``Geom-1'' and ``Geom-2'' denote the two geometries analyzed in Sec.~\ref{sec:results}. (b) O-grid mesh around an airfoil. (c) Near-wall mesh detail.}
	\label{fig:airfoil_dataset}
\end{figure}

For Q5, we consider flow over Joukowsky airfoils.
The velocity field is denoted by $\bm u(\bm x)$, where $\bm x=(x,y)$ is the spatial coordinate and $\rho$ is the density. 
The airfoil geometry is generated by the Joukowsky transformation
\begin{equation}
	z=\zeta+\frac{1}{\zeta},
	\label{eq:joukowsky}
\end{equation}
where the circle radius $R$ and the center offset $\alpha+i\beta$ control the airfoil thickness and camber. 
The angle of attack is denoted by ${\rm AoA}$ and the chord length by $c$. 
The training set contains symmetric airfoils with $\beta^\ast=0$ and ${\rm AoA}\in[0^\circ,10^\circ]$, while the OOD tests include larger angles of attack and a cambered geometry with $\beta^\ast\ne0$.

For Q5, reference solutions are generated with the finite-volume CFD solver OpenCFD-EC \cite{qi2021direct,men2023direct}. 
The freestream Mach number is fixed at $M_\infty=0.15$, corresponding to a low-Mach, nearly incompressible flow regime. 
The computational domain is
\begin{equation*}
	\Omega=\{(x,y)\in\mathbb{R}^2\,|\,x^2+y^2<(25c)^2\},
\end{equation*} 
and each computational domain is discretized by an O-grid with resolution $96\times128$, as illustrated in Fig.~\ref{fig:airfoil_dataset}. 
The governing equations are discretized by a cell-centered finite-volume method (FVM). 
Convective fluxes are evaluated using Roe flux-difference splitting, with third-order upwind reconstruction of the conservative variables at cell faces. 
Time advancement is performed using an implicit lower--upper symmetric Gauss--Seidel (LU-SGS) scheme until a steady solution is obtained. 

All quantities marked by a superscript asterisk $(\cdot)^*$ are nondimensionalized by their corresponding reference scales, such as $M$, $L$, $K$, $G$, $\Gamma$, $\mu_0$, and $\tau$. 
The reference scales are defined as follows. $M$ is the reference mass, $L$ is the reference length or displacement scale, and $\tau$ is the reference time. 
The derived scales are $K=M/\tau^2$ for spring stiffness, $G=L/\tau^2$ for gravitational acceleration, $\mu_0=M/\tau$ for viscous damping, $ML/\tau$ for momentum, and $\Gamma_0=L^2/\tau$ for point-vortex circulation. 
For Q5, the geometric quantities $\alpha,\beta,R,x,y$ are nondimensionalized by the chord-based length scale $L=c$, velocities by the freestream speed $U_\infty$, and pressure by $\rho U_\infty^2$.
In the reported nondimensional datasets these reference values are set to unity unless otherwise stated.
The training domains, action functionals, and test conditions are summarized in Table~\ref{tab:benchmark_setup}; $\underline{\mathbf{bold}}$ text indicates temporal, parameter, or geometric extrapolation outside the training domain.

\subsection{Score model architecture and training}

\begin{figure}
	\centering
	\includegraphics[width=1.0\linewidth]{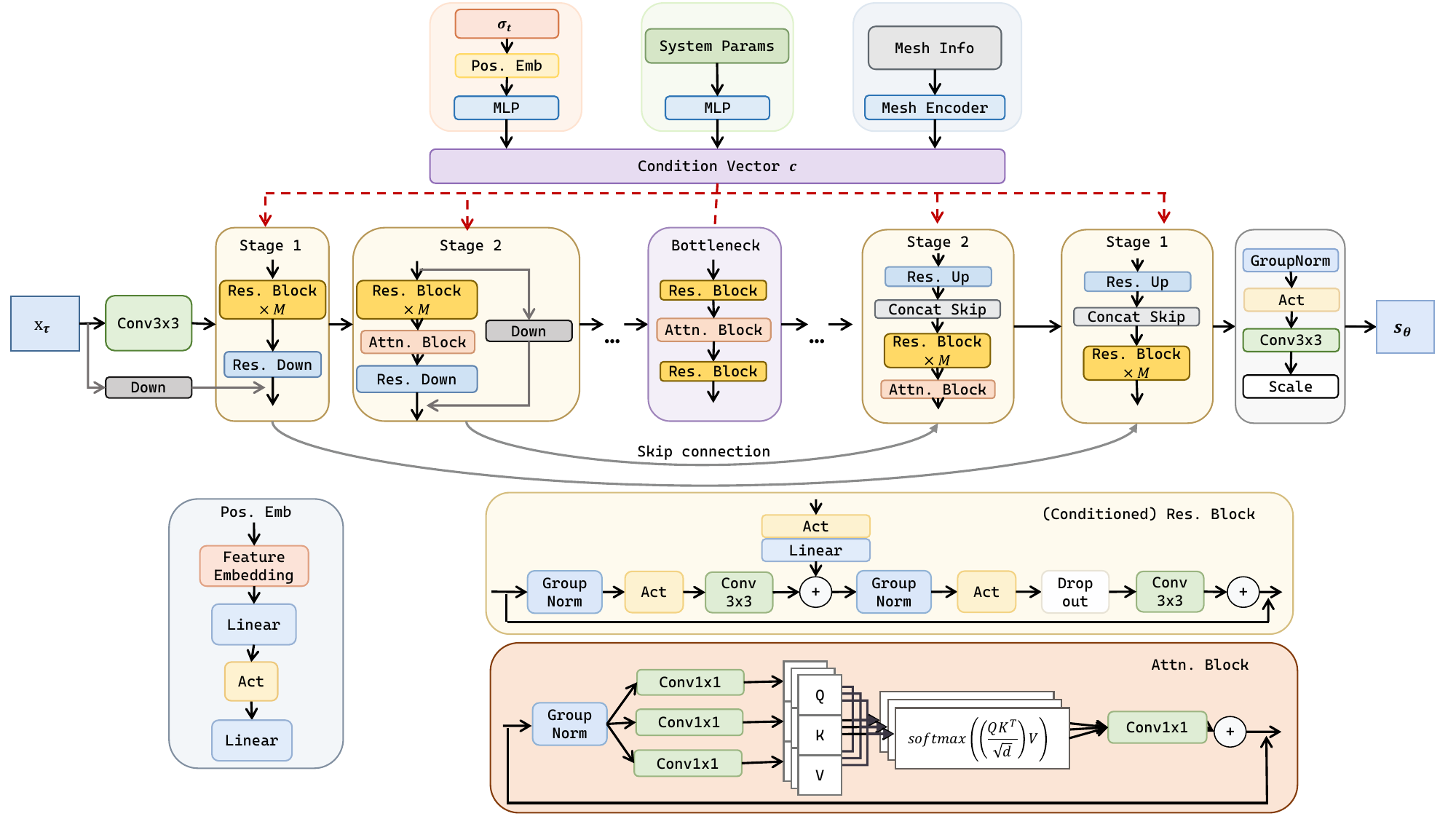}
	\caption{Conditional score-model architecture. The noisy state, noise-level embedding, physical-condition embedding, and, for Q5, mesh embedding are combined in a U-Net backbone with residual blocks, skip connections, and attention modules.}
	\label{fig:score_network}
\end{figure}

The conditional score function $S_\theta({\bf X}_\tau,\sigma_\tau;\bm c)$ is represented by a conditional U-Net, as shown in Fig.~\ref{fig:score_network}. 
The network takes the noisy state ${\bf X}_\tau$, the noise level $\sigma_\tau$, and the condition vector $\bm c$ as inputs, and outputs a score estimate with the same dimension as ${\bf X}_\tau$. 
For Q1--Q4, the input is a one-dimensional discretized trajectory in phase space. 
For Q5, the input is a two-dimensional flow field on the O-grid.

The architecture consists of a multi-scale encoder--decoder with skip connections. 
Each resolution level contains residual blocks, and self-attention is applied at selected resolutions to capture long-range correlations along the trajectory or over the flow field. 
The noise level $\sigma_\tau$ is embedded through a positional embedding, while the physical condition vector $\bm c$ is processed by a multilayer perceptron. 
For the airfoil problem, the mesh is additionally encoded and used as part of the conditioning information. 
The resulting conditioning embeddings are injected into the residual blocks as feature-wise biases. 
This design allows a single score model to represent a family of solutions over the training condition domain.
The case-dependent input dimensions, resolutions, conditioning variables, training parameters, SDE settings, and physical-refinement settings are summarized in Table~\ref{tab:score_lapg_setup}.

\begin{table*}[t]
	\centering
	\caption{Score-model training and LAPG inference settings. The same conditional U-Net design is used for all cases, with case-dependent input size, resolution, and condition variables.}
	\label{tab:score_lapg_setup}
	\scriptsize
	\setlength{\tabcolsep}{5pt}
	\renewcommand{\arraystretch}{1.2}
	\begin{tabular}{llccccc}
		\hline
		\multicolumn{2}{c}{Setting} & Q1 & Q2 & Q3 & Q4 & Q5 \\ \hline
		
		\multirow{4}{*}{Data structure}
		& Number of samples & 1000 & 8000 & 8000 & 2000 & 5115 \\
		& Batch size & 64 & 64 & 64 & 64 & 16 \\
		& Resolution & 128 & 128 & 2048 & 128 & $96\times128$ \\
		& Input channels & 2 & 2 & 2 & 4 & 3 \\ \hline
		
		\multirow{4}{*}{Model architecture}
		& Residual blocks, $M$ & 1 & 1 & 1 & 1 & 1 \\
		& Attention resolution & 64 & 64 & 1024 & 64 & $(48,64)$ \\
		& Channel multipliers & $(1,2,2,2)$ & $(1,2,2,2)$ & $(1,2,2,2,2,2,2)$ & $(1,2,2,2)$ & $(1,2,2,2,2)$ \\
		& Conditioning variables & $g$ & $q_0,p_0,k$ & $q_0,k,\mu$ & $x_{2,0},\Gamma_2$ & $\alpha,R,{\rm AoA}$, mesh \\ \hline
		
		\multirow{5}{*}{Training process}
		& Optimizer & \multicolumn{5}{c}{Adam} \\
		& Learning rate & \multicolumn{5}{c}{$2.0\times10^{-4}$} \\
		& Betas & \multicolumn{5}{c}{$(0.9,0.999)$} \\
		& Eps & \multicolumn{5}{c}{$1.0\times10^{-8}$} \\
		& EMA rate & \multicolumn{5}{c}{$0.999$} \\ \hline
		
		\multirow{3}{*}{SDE}
		& $\sigma_{\min}$ & \multicolumn{5}{c}{$0.01$} \\
		& $\sigma_{\max}$ & \multicolumn{5}{c}{$50.0$} \\
		& Sampling steps & \multicolumn{5}{c}{$1000$} \\ \hline
		
		\multirow{6}{*}{LAPG guidance}
		& Perturbation magnitude $\epsilon$ & $10^{-12}$ & $10^{-12}$ & $10^{-12}$ & $10^{-12}$ & $5\times10^{-4}$ \\
		& Perturbation function &  $\sin$--$\cos$ & $\sin$--$\sin$ & $\sin$--$\cos$ & $\sin$--$\cos$ & $\mathcal{N}(0,1)$ \\
		& Perturbation directions & $126$ & $126$ & $2046$ & $126$ & $10$ \\
		& Physical optimizer & Adam & Adam & SGDM & SGDM & SGDM \\
		& Max. refinement iterations & $4\times 10^4$ & $2\times 10^4$ & $2.5\times 10^5$ & $1.2\times 10^5$ & $5\times 10^3$\\
		& Refinement learning rate & $4 \times 10^{-1}$ & $1\times 10^{-2}$ & $ 2 \times 10^{-2}$ & $4\times 10^{-1}$ & $4\times 10^{-5}$  \\  \hline
		
	\end{tabular}
\end{table*}

\subsection{LAPG guidance perturbations}

For Q1--Q4, the LAPG physical score is evaluated on the full generated phase-space trajectory ${\bf X}=\{\bm x_i\}_{i=0}^{N}$, where $\bm x_i=[\bm q_i,\bm p_i]^T$. 
The virtual perturbation in Eq.~\eqref{eq:perturbation} is applied componentwise to the trajectory. 
Following the endpoint condition of the variational principle, the coordinate perturbations are chosen as sine modes
\begin{equation}
	(\xi_q^{(r)})_i
	=
	s_q^{(r)}
	\sin\left(\frac{r\pi i}{N}\right),
	\qquad
	r=1,\ldots,R,\quad i=0,\ldots,N,
	\label{eq:xi_q}
\end{equation}
where $s_q^{(r)}=\pm1$ is a random sign and each mode vanishes at both endpoints. 
The momentum perturbations are not constrained by the endpoint condition and are selected according to the system:
\begin{equation}
	(\xi_p^{(r)})_i=
	\begin{cases}
		s_p^{(r)}\cos(r\pi i/N), & \text{Q1, Q3, Q4},\\
		s_p^{(r)}\sin(r\pi i/N), & \text{Q2},
	\end{cases}
	\qquad r=1,\ldots,R,
	\label{eq:xi_p}
\end{equation}
with $s_p^{(r)}=\pm1$. 
The number of perturbation functions per generalized-coordinate or momentum component is $R=126$ for Q1, Q2, and Q4, and $R=2046$ for Q3. 
Thus, the notation ``$\sin$--$\cos$'' in Table~\ref{tab:score_lapg_setup} denotes a set of sine perturbations for generalized coordinates and cosine perturbations for momenta, rather than a single perturbation function.

For Q5, the physical refinement is applied to the circulation degree of freedom in the potential-flow representation. 
The complex potential in the $\zeta$-plane is written as
\begin{equation}
	F(\zeta)
	=
	A e^{-i\theta}\zeta
	+
	\frac{B e^{i\theta}}{\zeta}
	+
	\frac{i\Gamma}{2\pi}\log\zeta ,
	\label{eq:complex_potential}
\end{equation}
where $A$ and $B$ are determined by the freestream and no-penetration boundary conditions, and $\Gamma$ controls the circulation. 
Here $\theta$ denotes the freestream angle of attack. 
For a circle of radius $R$ in the $\zeta$-plane, written in coordinates centered at the circle center $\zeta_c=\alpha+i\beta$, the standard Joukowski potential gives $A=U_\infty$ and $B=U_\infty R^2$. 
We therefore perturb only $\Gamma$, using $\Gamma^{(r)}=\Gamma+\epsilon\xi_\Gamma^{(r)}$ with $\xi_\Gamma^{(r)}\sim\mathcal{N}(0,1)$ and $r=1,\ldots,10$. 
In Q5, the physical refinement is carried out on the circulation parameter $\Gamma$ in the auxiliary potential-flow representation, not by independently perturbing every velocity and pressure value on the CFD grid. Changing $\Gamma$ modifies the potential-flow velocity and pressure field through Eq.~\eqref{eq:complex_potential}; the action-like quantity is evaluated for 10 Gaussian circulation perturbations, and the mean squared finite-difference residual provides the guidance loss. Thus $\Gamma$ is the variational degree of freedom updated during the Q5 refinement, and the resulting circulation change alters the reconstructed airfoil flow field.
The perturbation magnitudes, perturbation functions, and number of finite-difference directions in the LAPG refinement stage are summarized in Table~\ref{tab:score_lapg_setup}.
After the multi-directional action-variation residual is evaluated, the gradient of $\log p_s$ is computed by automatic differentiation and used in the refinement stage.

\subsection{PINN baseline}

\begin{table*}[t]
	\centering
	\caption{PINN architectures and loss weights. The numbers in brackets indicate the number of neurons in each hidden layer. N/A denotes a loss component not used for the corresponding system.}
	\label{tab:pinn_setup}
	\scriptsize
	\setlength{\tabcolsep}{4pt}
	\renewcommand{\arraystretch}{1.2}
	\begin{tabular}{p{0.7cm}p{3.4cm}p{2.5cm}p{1.5cm}p{1.1cm}p{1.1cm}p{1.1cm}p{1.1cm}p{1.1cm}}
		\hline
		ID & Architecture & Input & Output & $\lambda_{\rm phys}$ & $\lambda_{\rm ic}$ & $\lambda_{\rm bc}$ & $\lambda_{\rm data}$ & LR \\ \hline
		Q1 & $(2,20,20,20,2)$ & $t,g$ & $h,p$ & 1 & 1 & N/A & 1 & $10^{-3}$ \\
		Q2 & $(5,32,64,64,32,2)$ & $t,q_0,p_0,k,m$ & $q,p$ & 1 & 1 & N/A & 1 & $10^{-3}$ \\
		Q3 & $(5,32,64,64,32,2)$ & $t,q_0,k,\mu,m$ & $q,p$ & 1 & 1 & N/A & 1 & $10^{-3}$ \\
		Q4 & $(3,32,64,64,32,4)$ & $t,x_{2,0},\Gamma_2$ & $x_1,y_1,x_2,y_2$ & 1 & 1 & N/A & 1 & $10^{-3}$ \\
		Q5 & \makecell[l]{$(6,128,128,128,128,128,$\\$128,128,128,128,3)$} & \makecell[l]{$x,y,{\rm AoA},$\\$\alpha,\beta,R$} & $u,v,p$ & 1 & N/A & $2\times10^3$ & $2\times10^4$ & $2\times10^{-4}$ \\ \hline
	\end{tabular}
\end{table*}

We compare LAPG with a training-time physics-constrained PINN baseline. 
For each system, the PINN directly approximates the solution map
$\hat{\bm y}_\theta(\bm z,\bm c)$, where $\bm z$ denotes the independent coordinate variables, such as $t$ for ODE systems and $(x,y)$ for the airfoil-flow problem. 
The condition vector $\bm c$ contains the corresponding physical parameters, initial conditions, or geometric parameters. 
For Q1--Q4, the PINN is a fully connected multilayer perceptron with hyperbolic-tangent activation. 
For Q5, a deeper fully connected network with Swish activation is used to represent the velocity and pressure fields. 
All derivatives required in the governing-equation residuals are computed by automatic differentiation.

The PINN baseline is chosen as a training-time physics-informed competitor. 
The first four benchmarks are smooth low-dimensional systems with known governing equations, a setting favorable to fully connected PINNs. 
For Q5, the baseline is a deeper network trained with data, boundary, and physics losses, and related work has shown that neural networks and data-assisted PINNs can accurately predict airfoil flow fields within the training parameter range \cite{bhatnagar2019prediction,harmening2024data}. 
The comparison therefore targets the distinction between training-time physical regularization and inference-time physical guidance.

The PINN is trained by minimizing a composite loss,
\begin{equation}
	\mathcal{L}_{\rm PINN}
	=
	\lambda_{\rm phys}\mathcal{L}_{\rm phys}
	+
	\lambda_{\rm ic}\mathcal{L}_{\rm ic}
	+
	\lambda_{\rm bc}\mathcal{L}_{\rm bc}
	+
	\lambda_{\rm data}\mathcal{L}_{\rm data},
	\label{eq:pinn_loss}
\end{equation}
where $\mathcal{L}_{\rm phys}$ penalizes the governing-equation residual, $\mathcal{L}_{\rm ic}$ enforces initial conditions, $\mathcal{L}_{\rm bc}$ enforces boundary conditions, and $\mathcal{L}_{\rm data}$ enforces agreement with the training data. 
The PINN uses the same training data as the score model. 
The architectures and loss weights are summarized in Table~\ref{tab:pinn_setup}.

\section{Results}
\label{sec:results}

We first evaluate LAPG on the four trajectory-generation benchmarks Q1--Q4. 
These systems test different extrapolation modes: long-time prediction beyond the training horizon and parameter shifts outside the training range. 
For Q1, the time horizon is doubled from $t^\ast\in[0,2]$ to $[0,4]$, while gravity is tested at $g^\ast=2$ and $30$ outside the training interval $[5,15]$. 
For Q2, trajectories are extended from one period to two periods, with stiffness tested at $k^\ast=3$ above the training interval $[0.5,1.5]$ and mass tested at $m^\ast=2$ beyond the fixed training value. 
For Q3, the damping coefficient is tested at $\mu^\ast=6$, twice the upper training limit, and additional cases combine out-of-range mass, stiffness, and damping. 
For Q4, the initial vortex separation and circulation are tested at $x_{2,0}^\ast=4$ and $\Gamma_2^\ast=4$, both well beyond their training intervals. All test conditions are summarized in Table~\ref{tab:benchmark_setup}.

\begin{figure}
	\centering
	\includegraphics[width=0.75\linewidth]{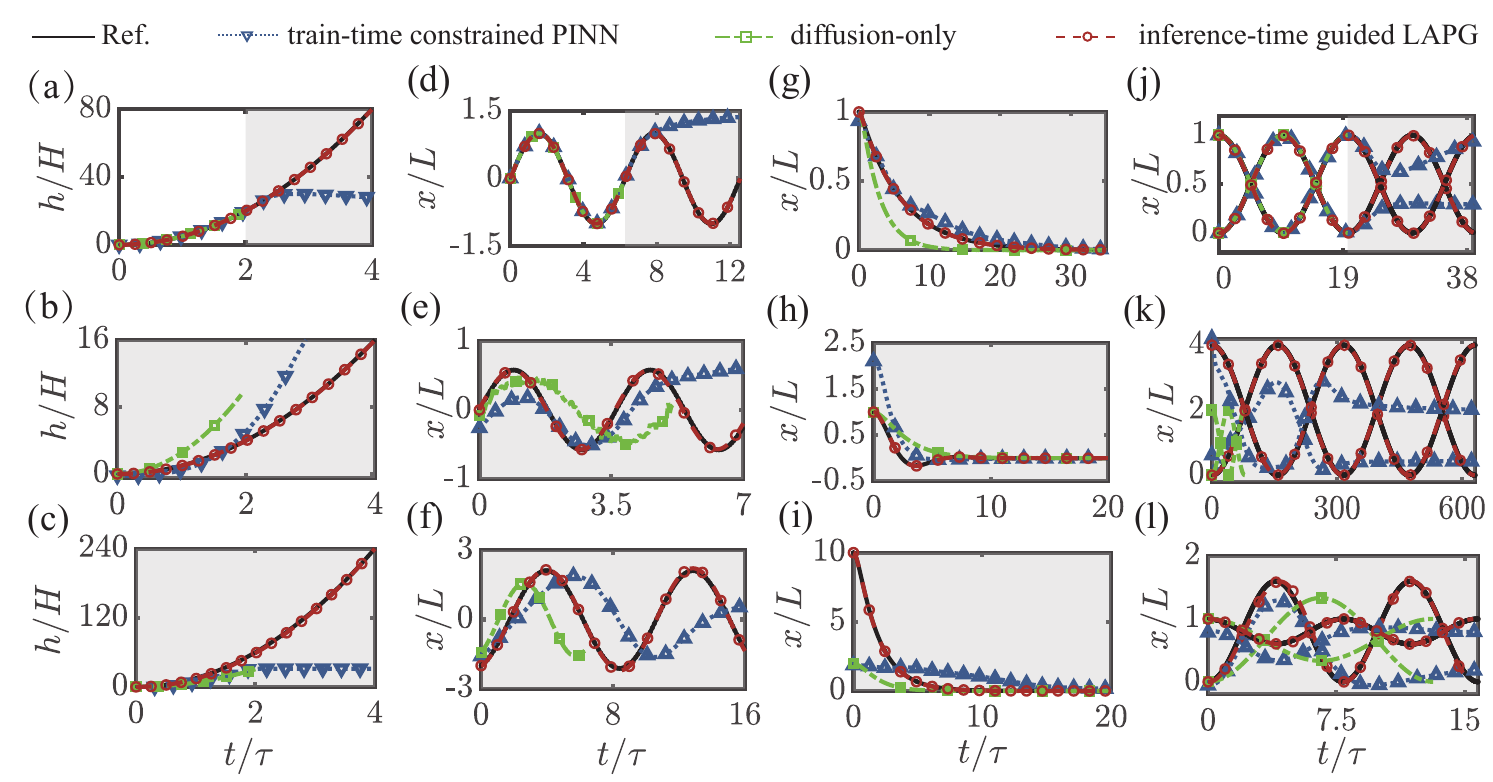}
	\caption{Trajectory generation in temporal and parameter extrapolation. Black: reference; green: data-driven diffusion-only; blue: training-time physics-constrained PINN; red: inference-time guided LAPG. Panels (a--c) Q1 (free fall); (d--f) Q2 (undamped spring--mass); (g--i) Q3 (damped spring--mass); (j--l) Q4 (point-vortex motion), with all trajectories plotted versus time $t/\tau$. Gray shading denotes extrapolation (either beyond the training horizon or outside the training parameter range); panels with a white-to-gray split indicate the end of the training time window.}
	\vspace{-5mm}
	\label{fig:ODEresults}
\end{figure}

\begin{figure}
	\centering
	\includegraphics[width=0.85\linewidth]{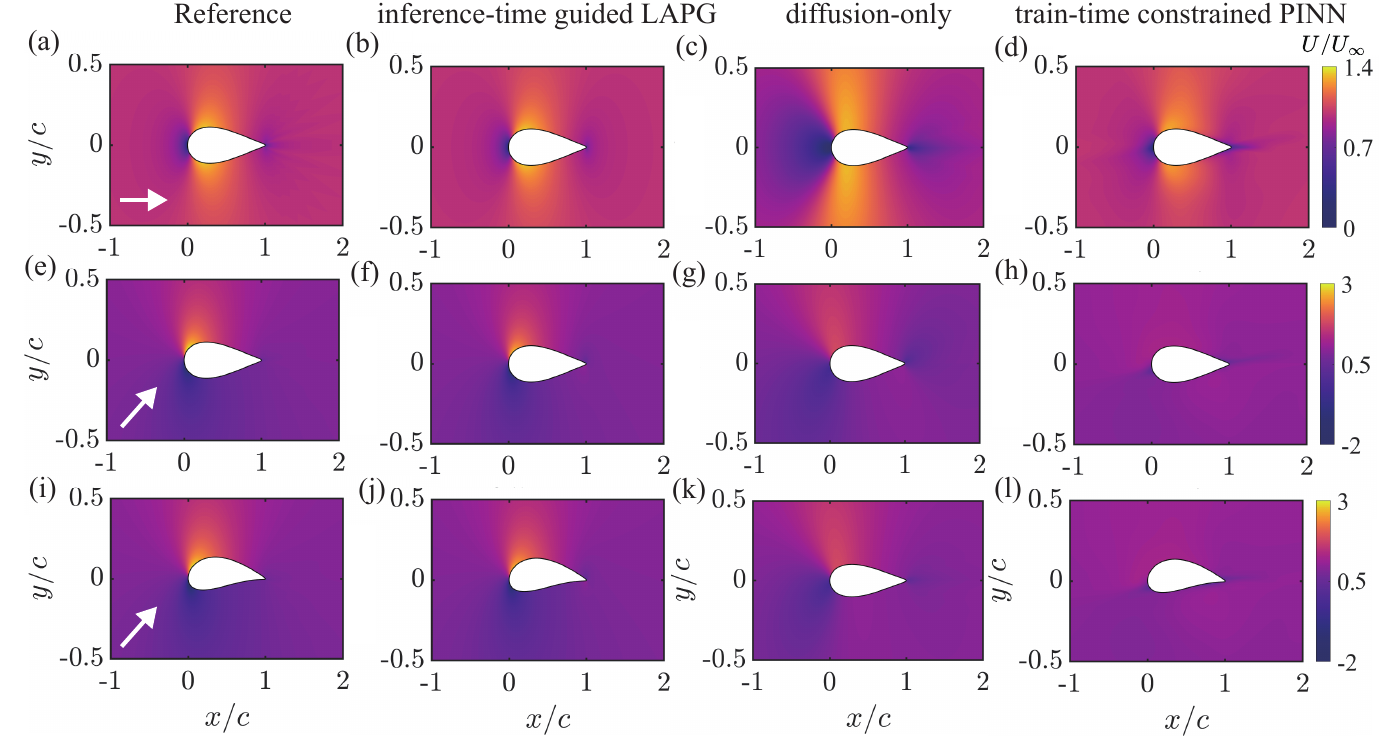}
	\caption{Potential flow past airfoils (Q5). Rows show one in-distribution target case (a--d) and two OOD target cases (e--l). The diffusion-only column (c,g,k) is generated at the nearest in-distribution condition $\bm c'$ before action-guided refinement, while the other columns correspond to the target condition $\bm c$. The colormap shows the streamwise velocity $U/U_\infty$, and the arrows indicate the freestream direction. Panels (a,e,i) reference; (b,f,j) LAPG; (c,g,k) diffusion-only; (d,h,l) PINN.}
	\vspace{-5mm}
	\label{fig:PDEresults}
\end{figure}

To make the role of each stage explicit, Figs.~\ref{fig:ODEresults} and~\ref{fig:PDEresults} include the data-driven diffusion-only output in addition to the PINN and LAPG results. The diffusion-only result is the nearest-condition score-model proposal generated before the action-derived refinement is activated. It is therefore not a separately trained target-condition solver, but a direct visualization of what the learned generator supplies before inference-time physics guidance.

Figure~\ref{fig:ODEresults} compares the generated trajectories with the reference solutions, the diffusion-only sample, the training-time physics-constrained PINN baseline, and the LAPG result.
For the in-distribution portions of the trajectories, shown by the white background in Fig.~\ref{fig:ODEresults}, both LAPG and PINN reproduce the reference solutions accurately. 
The main differences become pronounced in the extrapolation regime, shown by the gray shaded regions. 
In Q1, LAPG preserves the correct free-fall trend when the time horizon is extended and when gravity is shifted outside the training domain. 
In Q2, LAPG remains phase-accurate over two periods and maintains the correct amplitude under extrapolated mass or spring stiffness. 
By contrast, the PINN baseline develops visible phase drift and amplitude errors as the extrapolation distance increases.

The damped spring--mass system Q3 provides a more stringent test because the solution must capture both oscillatory motion and dissipative decay. 
LAPG accurately follows the decay envelope under extrapolated damping and remains stable under simultaneous shifts in mass, stiffness, and damping. 
For the point-vortex system Q4, LAPG captures the long-time vortex motion and remains accurate when either the initial separation or the circulation ratio lies outside the training range. 
In both cases, the action refinement changes the nearest-condition diffusion sample toward the target dynamics rather than only smoothing the trajectory pointwise.

We next evaluate Q5, potential flow over Joukowsky airfoils. 
This problem extends the evaluation from phase-space trajectories to spatially distributed fields. 
It also imposes a stronger OOD shift: the angle of attack is increased from the training range $[0^\circ,10^\circ]$ to $30^\circ$, and one test case moves from the symmetric training set with $\beta^\ast=0$ to a cambered geometry with $\beta^\ast=0.1$. 
Thus, the test probes both condition and geometry extrapolation.

Figure~\ref{fig:PDEresults} shows the streamwise velocity field $U/U_\infty$ for one in-distribution case and two OOD cases, comparing the reference solution, LAPG, diffusion-only output, and training-time physics-constrained PINN.
In the in-distribution case, LAPG and PINN recover the reference field, and the lift coefficients remain close to zero, as expected for a symmetric airfoil at zero angle of attack. 
The corresponding lift coefficients for the reference, LAPG, diffusion-only, and PINN fields are $C_l=(0.00,0.00,-0.02,0.06)$.
For the OOD cases, LAPG continues to reproduce the dominant flow features, including the leading-edge acceleration and the asymmetric velocity distribution associated with lift. 
The diffusion-only field is the nearest-condition data-driven proposal before refinement. Since it is generated at $\bm c'$ rather than directly constrained by the target condition $\bm c$, it captures an in-distribution flow pattern but does not impose the OOD angle of attack or geometry.
The PINN prediction is noticeably more diffusive and also underestimates the high-velocity region near the leading edge. 
This difference is reflected in the lift coefficient. 
For the two OOD cases, the reference, LAPG, diffusion-only, and PINN values are
\[
\begin{aligned}
	C_l &= (3.20,3.00,1.30,0.32),\\
	C_l &= (3.60,3.50,1.25,0.74),
\end{aligned}
\]
respectively.
Thus, while the PINN field may remain smooth, it does not recover the correct aerodynamic response under large extrapolation.

\begin{figure}
	\centering
	\includegraphics[width=0.70\linewidth]{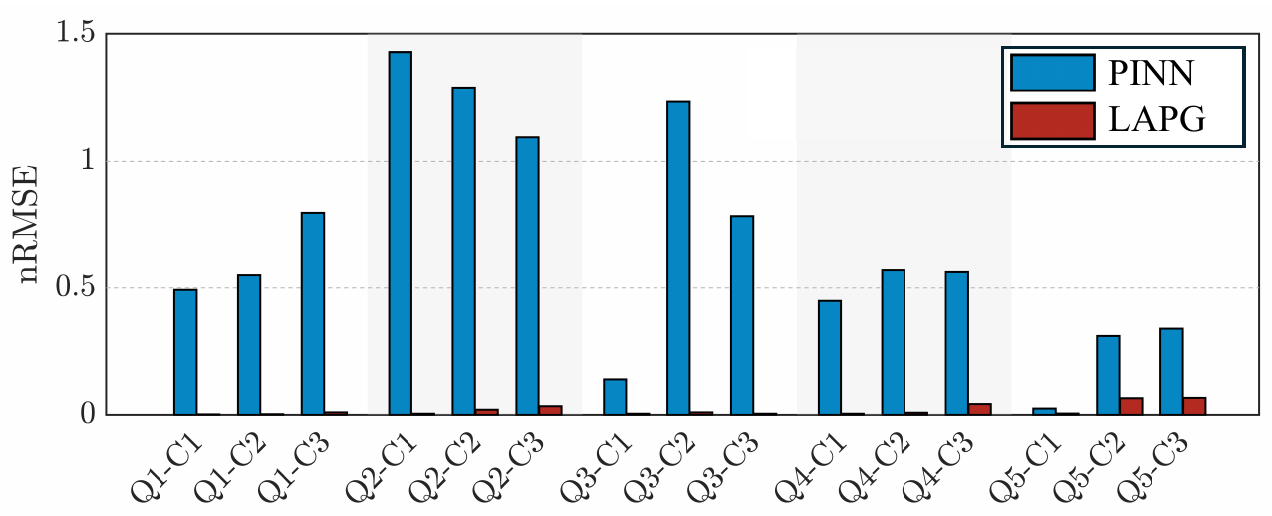}
	\caption{Quantitative error across all test cases listed in Table~\ref{tab:benchmark_setup}. Bars show the normalized root-mean-square error (nRMSE) for LAPG and the PINN baseline.}
	\label{fig:error_stat}
\end{figure}

Figure~\ref{fig:error_stat} summarizes the quantitative errors across all test cases. 
We use the normalized root-mean-square error as the metric:
\begin{equation}
	{\rm nRMSE}
	=
	\left(
	\frac{\sum_i \|\hat{\bm y}_i-\bm y_i^{\rm ref}\|_2^2}
	{\sum_i \|\bm y_i^{\rm ref}\|_2^2}
	\right)^{1/2},
	\label{eq:nrmse}
\end{equation}
where $\hat{\bm y}$ is the prediction and $\bm y^{\rm ref}$ is the reference solution. 
For Q1--Q4, the nRMSE is computed on the generated trajectories after time reconstruction. 
For Q5, the nRMSE is computed over the spatial grid of the velocity field.

Across the trajectory benchmarks, LAPG maintains low error in both temporal and parameter extrapolation, while the PINN error increases substantially in the OOD cases. 
The largest differences occur in cases where extrapolation changes the qualitative behavior of the solution, such as the oscillation period, decay rate, vortex orbit, or aerodynamic lift. 
These regimes are difficult for a fixed training-time model because the prediction is governed by extrapolation of the learned map. 
LAPG, in contrast, uses the learned score only to obtain a plausible proposal and then actively enforces the target physics through the action-derived score.

The quantitative results support the central mechanism of LAPG: the physical guidance term reduces the action variation during inference. 
Because the guidance is derived from a scalar variational functional, it acts on the full trajectory or field rather than on isolated output points. 
This global correction helps reduce accumulated phase error in long-time dynamics and improves parameter and geometry extrapolation. 
The results therefore indicate that inference-time variational guidance provides a practical route to improving the physical reliability of diffusion-based generators outside the training domain.

\section{Conclusion and Discussion}
\label{sec:conclusion}

LAPG improves extrapolation in the five benchmark problems considered here. The gains are clearest when the test condition changes a physical quantity that accumulates error over time or strongly affects the predicted results, such as phase, damping rate, vortex orbit, or lift.
The method refines the generated trajectory or field with an action-residual loss after diffusion sampling, making the physical constraint active at test time instead of relying only on constraints used during training.

Across representative ODE and PDE systems, including free fall, conservative and dissipative spring--mass dynamics, point-vortex motion, and potential flow over airfoils, LAPG improves temporal, parameter, and geometric extrapolation. 
Compared with training-time physics-informed baselines, the method better preserves phase accuracy, dissipative decay, vortex motion, and aerodynamic response in OOD regimes. 
These results indicate that inference-time variational guidance can improve the reliability of diffusion-based physical generators when the learned score alone would otherwise be evaluated beyond its training support.

Compared with existing physics-informed methods, an advantage of the proposed formulation is that the physical constraint enters through a global scalar action functional evaluated on the full trajectory or field. 
This differs from residual-based training objectives, where equation residuals, initial conditions, boundary conditions, and data losses often require problem-dependent weight balancing. 
By constructing a single action-based physical prior, LAPG avoids much of this multi-term loss-balancing burden while retaining a differentiable mechanism for enforcing physical consistency during sampling.

The framework also has limitations. 
It requires an appropriate action or action-like variational functional for the physical system of interest, which may be difficult to identify for complex dissipative, turbulent, multiphysics, or strongly constrained systems. 
The inference-time refinement also introduces additional computational cost because the action variation and its gradient must be evaluated during sampling. 
Moreover, the final result can depend on the diffusion-generated proposal, the perturbation used to estimate the action variation, and the optimization parameters in the refinement stage.
Future work will focus on combining global action-based guidance with complementary local constraints when the variational functional alone does not encode all relevant physical requirements. 

\bibliographystyle{unsrtnat}
\bibliography{references}

\end{document}